# Efficient Sequential Neural Network with Spatial-Temporal Attention and Linear LSTM for Robust Lane Detection Using Multi-Frame Images

Sandeep Patil[#], Yongqi Dong[#], Haneen Farah, and Hans Hellendoorn

*Abstract*—**Lane detection is a crucial perception task for all levels of automated vehicles (AVs) and Advanced Driver Assistance Systems, particularly in mixed-traffic environments where AVs must interact with human-driven vehicles (HDVs) and challenging traffic scenarios. Current methods lack versatility in delivering accurate, robust, and real-time compatible lane detection, especially vision-based methods often neglect critical regions of the image and their spatial-temporal (ST) salience, leading to poor performance in difficult circumstances such as serious occlusion and dazzle lighting. This study introduces a novel sequential neural network model with a spatial-temporal attention mechanism to focus on key features of lane lines and exploit salient ST correlations among continuous image frames. The proposed model, built on a standard encoder-decoder structure and common neural network backbones, is trained and evaluated on three large-scale open-source datasets. Extensive experiments demonstrate the strength and robustness of the proposed model, outperforming state-of-the-art methods in various testing scenarios. Furthermore, with the ST attention mechanism, the developed sequential neural network models exhibit fewer parameters and reduced Multiply-Accumulate Operations (MACs) compared to baseline sequential models, highlighting their computational efficiency. Relevant data, code, and models are released at https://doi.org/10.4121/4619cab6-ae4a-40d5-af77-582a77f3d821.**

Manuscript submitted on 31 January, 2026. This work was supported by the Applied and Technical Sciences (TTW), a subdomain of the Dutch Institute for Scientific Research (NWO) through the Project Safe and Efficient Operation of Automated and Human-Driven Vehicles in Mixed Traffic (SAMEN) under Contract 17187. (# Sandeep Patil and Yongqi Dong are co-first authors.) (Corresponding author: Yongqi Dong.)

Sandeep Patil is with Capgemini Nederland B.V., 3543 KA Utrecht, the Netherlands, and the Faculty of Mechanical Engineering, Delft University of Technology, 2628 CN Delft, the Netherlands (e-mail: contact@sandeeppatil.info).

Yongqi Dong is with the Faculty of Civil Engineering and Geosciences, Delft University of Technology, 2628 CN Delft, the Netherlands, and RWTH Aachen University, 52074 Aachen, Germany (e-mail: yongqi.dong@rwth-aachen.de).

Haneen Farah is with the Faculty of Civil Engineering and Geosciences, Delft University of Technology, 2628 CN Delft, the Netherlands (e-mail: h.farah@tudelft.nl).

Hans Hellendoorn is with the Faculty of Mechanical Engineering, Delft University of Technology, 2628 CN Delft, the Netherlands (e-mail: j.hellendoorn@tudelft.nl).

## I. INTRODUCTION

AUTOMATED vehicles (AVs) are increasingly expected to operate in mixed traffic environments, where they must safely and efficiently interact with human-driven vehicles (HDVs) and face challenging traffic scenarios [1], [2]. In such settings, robust perception of road geometry and traffic infrastructure is a prerequisite for safe navigation, decision-making, and motion planning. Among various perception tasks, lane detection is a crucial capability, as it enables vehicles to locate and position themselves within lanes by identifying and predicting lane markings, providing essential reference information for localization, lane keeping, trajectory planning, and interaction-aware control. While various sensors, e.g., mono-camera, stereo-camera, radar, and lidar, could be applied in the process of detecting the lane boundaries for accurate localization [3], [4], the most common, feasible, and successful approach is vision-based lane marking detection [5], [6].

Conventional vision-based methods usually handle lane detection utilizing specialized hand-crafted low-level features with traditional computer vision techniques, e.g., Inverse Perspective Mapping applied in the image pre-processing stage [7], [8]; Hough transform applied for feature extraction [9], [10]; Gaussian filters and Random Sample Consensus (RANSAC) employed in the post-processing process to smooth the lane detection results [11], [12]. These traditional methods suffer from many shortcomings, e.g., they require hand-crafted features which are always complex and time-consuming but not necessarily suitable or effective enough, and they usually use one single image to detect the lane, thus cannot handle some extremely challenging driving scenarios.

Recent advances in computational hardware, along with rapid developments in neural network models, have enabled deep learning (DL) based lane detection methods to extract useful features automatically. They have been widely used to eliminate intermediate feature crafting, as well as enable end-to-end lane detection, outperforming traditional approaches [13], [14]. Usually, deep Convolutional Neural Networks (CNNs) have been widely adopted for their superior abilities in image feature abstraction, demonstrating exceptional performance in lane detection tasks, e.g., in [13], [15]. In addition to CNNs, other architectures like Recurrent Neural Networks (RNNs), Generative Adversarial Networks (GANs), and Vision Transformers (ViTs) have also been explored.

RNNs, known for their capability to process sequential data, are adept at abstracting and predicting time-series features. Consequently, they are employed to model sequential patterns within a single image [16] or across frames in continuous images for lane detection [17], [18]. GANs, which leverage two neural networks competing in a shared task have been used for data augmentation (e.g., generating synthetic lane images) and transfer learning applications in lane detection [19]. Recently, ViTs, adapted from the original Transformer architecture renowned for its success in natural language processing tasks, have been applied to computer vision problems, including lane detection. Studies such as [20], [21] utilize the self-attention mechanism inherent in Transformers to focus on salient regions in an image, improving lane detection accuracy. However, these approaches predominantly rely on single-image, overlooking temporal correlations and varying importance of frames in continuous driving scenarios.

A few studies have attempted to combine CNNs and RNNs to detect lane markings through continuous driving scene image frames [17], [18], [22], [23]. However, these approaches fail to fully exploit the inherent properties of lanes and often overlook the salient spatial-temporal (ST) correlations and dependencies among critical regions across sequential frames. As a result, their performance remains unsatisfactory under highly challenging driving conditions.

To address the aforementioned research gaps and improve the performance of vision-based lane detection, this study introduces a novel efficient sequential neural network architecture with the proposed ST attention mechanisms. The developed model formulates lane detection as a segmentation task and takes multiple continuous image frames as input. By effectively extracting key features and leveraging salient correlations across these frames, the proposed approach strongly exploits the ST information inherent in the driving scene. Built on a standard encoder-decoder framework and utilizing labeled ground truth from the final image in the sequence, the model employs a supervised, end-to-end learning strategy. The primary contributions of this paper are: 1) Introduction of ST attention mechanisms: Three attention model variants are proposed and implemented to improve feature extraction; 2) Strong exploitation of ST correlations: The proposed ST attention effectively captures and utilizes salient ST relationships among different regions in continuous image frames; 3) Superior performance: Extensive experiments demonstrate that the proposed model outperforms state-of-the-art baselines in both normal and challenging driving scenarios; 4) Lightweight architecture: The proposed model is more compact compared to other sequential models designed for multi-frame input, making it computationally efficient; 5) Robustness to unseen data: Qualitative evaluations show that the model maintains high robustness on entirely new and unlabeled datasets, unseen during training.

## II. PROPOSED METHOD

### A. Overall Architecture Description

Inspired by the human visual attention mechanism and considering that traffic lanes are of long thin line structures with strong spatial correlation, for vision-based lane detection, certain regions of the images and certain frames in the continuous driving scenes deserve more attention than other areas and frames. Moreover, it is witnessed that fusing CNN and RNN with hybrid deep neural network (DNN) architectures can make use of multiple continuous image frames to further improve lane detection performance [17], [18], [22], [23]. With all these clues, this study develops a novel dedicated ST attention mechanism for the lane detection task to fill the aforementioned research gaps. With the proposed ST attention mechanism, three model variants are implemented under hybrid sequential end-to-end DNN structures fusing CNN-based encoder-decoder and temporal RNN (e.g., Long Short-term Memory (LSTM)). On the whole, regarding vision-based lane detection as a segmentation task, the proposed model adopts a sequence-to-one architecture, i.e., it takes a sequence of multiple continuous image frames as inputs and outputs the detection result of the very last image frame. The regular CNN-based encoder-decoder neural network, UNet [24], is served as the network backbone. In UNet, the encoder module and the decoder module both contain four convolutional blocks (in one of the implementations in this paper, the first block of the encoder was replaced with Spatial CNN (SCNN) [13], which is discussed later in the Ablation Study). The proposed attention module is embedded between the encoder and decoder. The encoder module extracts useful features from the input continuous frames and feeds them to the attention module for further ST feature integration. The attention module can detect salient ST relevances and dependencies among the extracted feature maps of the consecutive frames and pass these integrated features to the decoder. Lastly, the decoder module upsamples and decodes the integrated feature maps to the same size as the input image and outputs the detected lanes. Note that, in the adopted UNet backbone, similar to [24], a skip connection is applied between the encoder and decoder with concatenating operation so that the decoder can reuse the extracted features and retain information from the encoder.

An architecture overview of the proposed method is illustrated in **Fig. 1**, and detailed implementation is further elaborated in the following sections.

### B. Spatial-temporal (ST) Attention Mechanism

The proposed attention module is developed to mimic human visual cognitive attention which demonstrates the ability to focus on important parts and ignore minor parts. The attention module helps the DNN to learn to focus on salient regions of the input images by assigning weights to each image frame and particular regions of each frame. With the help of the embedded temporal feature extractor, e.g., LSTM or Gated Recurrent Unit (GRU), the attention module can also extract important temporal dependencies over the input consecutive image frames.

As illustrated in **Fig. 1**, the attention module is applied when the input image sequences are downsized and the features are extracted by a series of convolution layers in the encoder. The attention module integrates the extracted features from the encoder and the hidden outputs produced by the embedded temporal feature extractor, e.g., LSTM/GRU. The LSTM/GRUs' hidden outputs of the very last previous time step and the input feature maps at the current time step can be combined using a set of attention weights. Activation of these weights can then be obtained to learn which image frames and which specific regions are important for lane detection. The weighted sum of input feature maps highlights the salient features, which are then processed by the temporal feature extractor to produce the output at the current time step and the updated hidden state. All the attention weights can be trained simultaneously with other neural network layer weights using the backpropagation mechanism. Equations **(1)-(12)** denote the mathematical explanation for the attention mechanism.

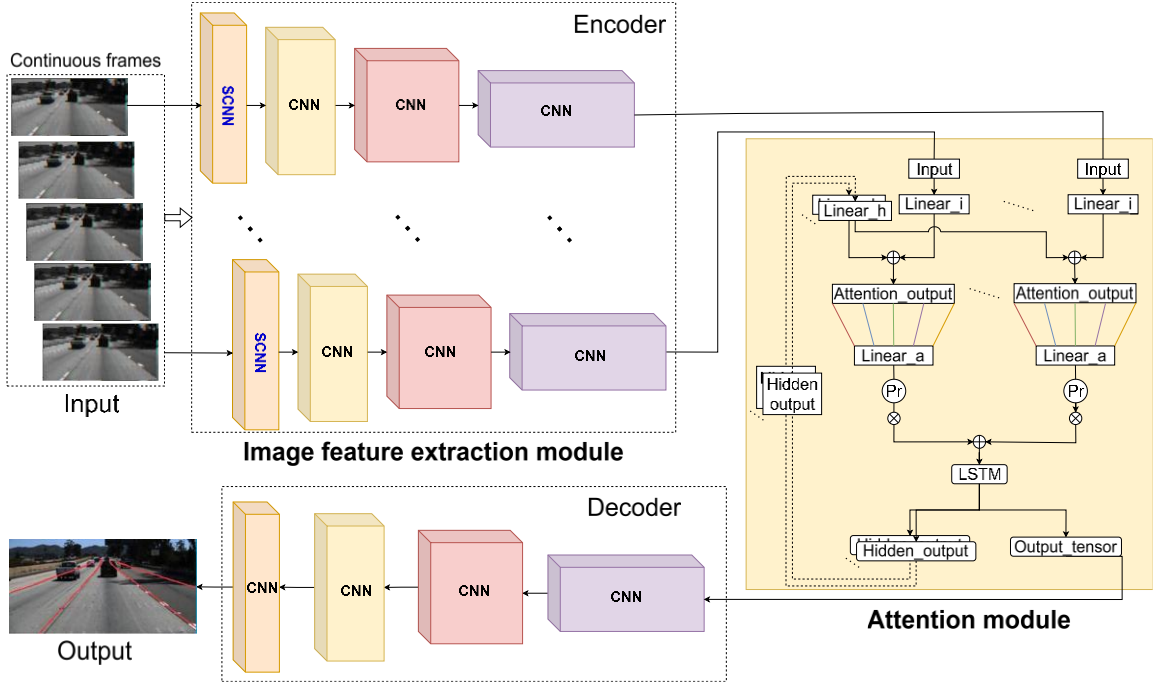

**Fig. 1.** The architecture of the proposed method.

The output of the final downsized convolutional block at time $t$ for the $n$-th frame (i.e., timestep $n$ within the image sequence) is denoted as $x_{down4}^{(t-N+n)}$, where $n = \{1, 2, \ldots, N\}$, and $N$ is the number of frames in the sequence, (in this implementation $N = 5$). The input sequence for the attention module is therefore defined as $\{x_{down4}^{(t-N+1)}, x_{down4}^{(t-N+2)}, \ldots, x_{down4}^{(t)}\}$. Please note there are two distinct temporal increments. The increment in $n$ corresponds to processing the subsequent image in the input sequence; where an increment in time $t$ reflects the real-world temporal progression, i.e., moving to the next input sequence. Then, within a certain selected sequence, the following computations are performed:

$$x^{(t-N+n)} = Conv(x_{down4}^{(t-N+n)}, k_{in}) \tag{1}$$
$$z^{(t-N+n)} = (U \odot x^{(t-N+n)}) + (H \odot h^{(t-N+n-1)}) \tag{2}$$
$$w^{(t-N+n)} = softmax(W \odot z^{(t-N+n)}) \tag{3}$$
$$\overline{x^{(t-N+n)}} = w^{(t-N+n)} \odot x^{(t-N+n)} \tag{4}$$

Here, "Conv" denotes the convolution operation, $\odot$ is the Hadamard (element-wise) multiplication, while "+" represents the element-wise addition operation. $k_{in}$ is a convolution layer with a kernel of size of $1 \times 1$ and 1 channel (as indicated by *In_Attention_Conv_5_1* in **TABLE I**). The matrices $U, H, W$ are the learnable weights that can be configured as trainable vectors of size $1 \times 1$ or $1 \times 128$, or as a trainable fully connected layer of size $1 \times 128$. $x^{(t-N+n)}$ and $z^{(t-N+n)}$ represent the intermediate outputs. $w^{(t-N+n)}$ denotes attention weights obtained from softmax operation, and $\overline{x^{(t-N+n)}}$ is the attention-based weighted output.

After processing the $N$ images, and getting the weighted output $\{\overline{x^{(t-N+1)}}, \overline{x^{(t-N+2)}}, \ldots, \overline{x^{(t)}}\}$ across the sequence, the following computations are carried out:

$$h^{(t)} = F(\{\overline{x^{(t-N+1)}}, \overline{x^{(t-N+2)}}, \ldots, \overline{x^{(t)}}\}, h^{(t-N+n-1)}) \tag{5}$$
$$x_{out} = Conv(h^{(t)}, k_{out}) \tag{6}$$

where $F$ stands for an embedded temporal feature extractor; $h^{(t)}$ is the hidden state vector initialized as $h^{(0)} = \mathbf{0}$ (zero vector) when $t = 0$ and $h^{(t)}$ will be updated with its new inheritor after the selected sequence is fully processed as in **equation (5)**; $h^{(t)}$ is also the output from F after processing N frames, i.e., $h^{(t-N+N)} = h^{(t)}$, which is then expanded to 512 channels by the *outconv* layer $k_{out}$; $k_{out}$ has a kernel size of $1 \times 1$ and 512 channels (indicated by *Out_Attention_Conv_5_2* in **TABLE I**); $x_{out}$ is the final output of the attention module which is then transferred to the decoder module.

The temporal feature extractor $F$ can be LSTM or GRU. Take LSTM for example, an LSTM unit is visualized in **Fig. 2**. Here, $C$ is the memory cell, while $i$, $f$, and $o$ stand for input gate, forget gate, and output gate, respectively, which regulate the flow of information. The key formulations of the LSTM are shown by **equations (7)-(12)**:

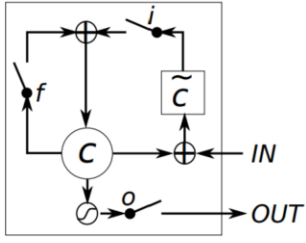

**Fig. 2.** An illustration of the LSTM unit [25].

$$f^{(t-N+n)} = \sigma\left(b^f + P^f\overline{x^{(t-N+n)}} + Q^f h^{(t-N+n-1)}\right) \quad (7)$$

$$i^{(t-N+n)} = \sigma\left(b^i + P^i\overline{x^{(t-N+n)}} + Q^i h^{(t-N+n-1)}\right) \quad (8)$$

$$\tilde{c}^{(t-N+n)} = g\left(b^c + P^c\overline{x^{(t-N+n)}} + Q^c h^{(t-N+n-1)}\right) \quad (9)$$

$$c^{(t-N+n)} = f^{(t-N+n)} \odot c^{(t-N+n-1)} + i^{(t-N+n)} \odot \tilde{c}^{(t-N+n)} \quad (10)$$

$$o^{(t-N+n)} = \sigma\left(b^o + P^o\overline{x^{(t-N+n)}} + Q^o h^{(t-N+n-1)}\right) \quad (11)$$

$$h^{(t-N+n)} = o^{(t-N+n)} \odot g(c^{(t-N+n)}) \quad (12)$$

where $g(\cdot)$ is the typically hyperbolic tangent function, and $\sigma$ is the activation function. $b^f$, $P^f$, $Q^f$ are biases, input weights, and recurrent weights for the forget gates; accordingly, $b^i$, $P^i$, $Q^i$ are those for the input gate; $b^c$, $P^c$, $Q^c$ are those for the current state of the memory cell; $b^o$, $P^o$, $Q^o$ are those for the output gate. At each time step, the current input vector $\overline{x^{(t-N+n)}}$ and the previous hidden state $h^{(t-N+n-1)}$ are combined and processed to produce the states of the forget gate $f^{(t-N+n)}$ (7), input gate $i^{(t-N+n)}$ (8), output gate $o^{(t-N+n)}$ (11), and the candidate memory update $\tilde{c}^{(t-N+n)}$ (9). The forget gate then determines which components of the previous cell state $c^{(t-N+n-1)}$ to retain, while the input gate modulates the contribution of the candidate cell state $\tilde{c}^{(t-N+n)}$, to yield the new cell state $c^{(t-N+n)}$ (10). Finally, the output gate filters the activated cell state through a nonlinear function $g(\cdot)$ to produce the hidden state $h^{(t-N+n)}$ (12), which is propagated as the recurrent input to the subsequent time step.

In the implementation, depending on different settings of the learnable weights $U, V, W$ in (2)-(3), three variants of the proposed attention module are developed and tested. They are temporal attention (Tem_Att, for short), spatial-temporal attention (ST_Att), and spatial-temporal attention model with fully connected layers (STFC_Att). The attention model is implemented after the encoder module (to be specific, the 4th down-sampling convolutional block, i.e., *Down_ConvBlock_4* in **TABLE I**) and before the decoder module (to be specific, the 1st upsampling convolutional block, i.e., *Up_ConvBlock_4* in **TABLE I**). One should notice that the attention model is modular in nature and can be adopted with any network backbone, not only UNet but also backbones such as SegNet [26] and fully convolutional networks [27].

### 1) Temporal attention

The design of the ST attention mechanism began with assessing the significance of each frame in a sequence for detecting lane markings in the current frame, which is implemented through the temporal attention (Tem_Att) module. As demonstrated in **Fig. 3(a)**, in the Tem_Att module, the learnable weights of $U$, $H$, and $W$ in **equations (2)-(3)** are trainable vectors and are illustrated by $V\_i$, $V\_h$, and $V\_a$ in **Fig. 3(a)** respectively. The three trainable vectors, each of size $1\times1$, dynamically adjust the contributions of input features, hidden state output, and the attention output based on the learned weights. The input features $x^{(t-N+n)}$ are modulated by the weight vector $V\_i$ and combined with the hidden output multiplied by $V\_h$ through element-wise addition to construct a summed intermediate attention signal $z^{(t-N+n)}$, as described in **equation (2)**. This attention signal is subsequently passed through a softmax activation function, shown as 'Pr' in **Fig. 3(a)**, to compute the attention weights $w^{(t-N+n)}$, as defined in **equation (3)**. These weights effectively prioritize the significance of each frame in the sequence.

Leveraging the LSTM unit (detailed in **equations (7)-(12)**), the hidden state $h^{(t)}$ contextualizes the input features by incorporating information from the entire sequence within the selected time window. The attention output $\overline{x^{(t-N+n)}}$ computed as a weighted combination of the input features $x^{(t-N+n)}$ and their respective attention weights $w^{(t-N+n)}$ (see **equation (4)**), captures these temporal dependencies. This output is processed through the LSTM and a convolutional layer (as outlined in **equation (6)**) to generate the module's final output $x_{out}$, which is subsequently passed to the decoder. When the three trainable vectors $V\_i$, $V\_h$, and $V\_a$ are of size $1\times1$, this approach ensures that the model dynamically adapts its focus to relevant temporal features in the image sequence.

### 2) Spatial-temporal attention

Observations showed that lane lines generally appear in specific regions within the image frames, with some features being more important than others. To address this, spatial attention is applied to each frame, and connecting with the temporal attention for the sequence of frames to form a spatial-temporal attention (ST_Att) module. This module uses three learnable weight vectors (size of $1 \times 128$), each multiplied by the input feature matrix, the previous hidden output, and the current step's attention output. The ST_Att module learns the importance of each feature without considering neighboring pixels (addressed by STFC_Att later). **Fig. 3(b)** illustrates the ST_Att structure, which functions similarly to Tem_Att, with different colored lines indicating connections between inputs, hidden outputs, and attention outputs with their weight matrices. Like Tem_Att, the attention output is scaled between 0 and 1 using the softmax function ('Pr' in **Fig. 3(b)**) and is then passed through a convolutional layer to the decoder. This mechanism ensures that the model emphasizes the most critical spatial features in each frame. When combined with the temporal modeling capabilities of the LSTM, it effectively leverages ST information across image frames in the sequence.

### 3) Spatial-temporal attention with fully connected layers

The spatial-temporal attention with fully connected layers (STFC_Att) module builds upon the ST_Att module by incorporating a fully connected mechanism to enhance feature learning. Unlike the one-to-one connections in ST_Att, the STFC_Att module employs many-to-many connections, where each learnable weight matrix is multiplied with all values of the input feature map, as illustrated by different color lines in **Fig. 3(c)**. This many-to-many connection allows the model to extract spatial dependencies between feature maps within the same image frame while concurrently capturing temporal features and correlations across consecutive frames with the assistance of the LSTM's hidden outputs.

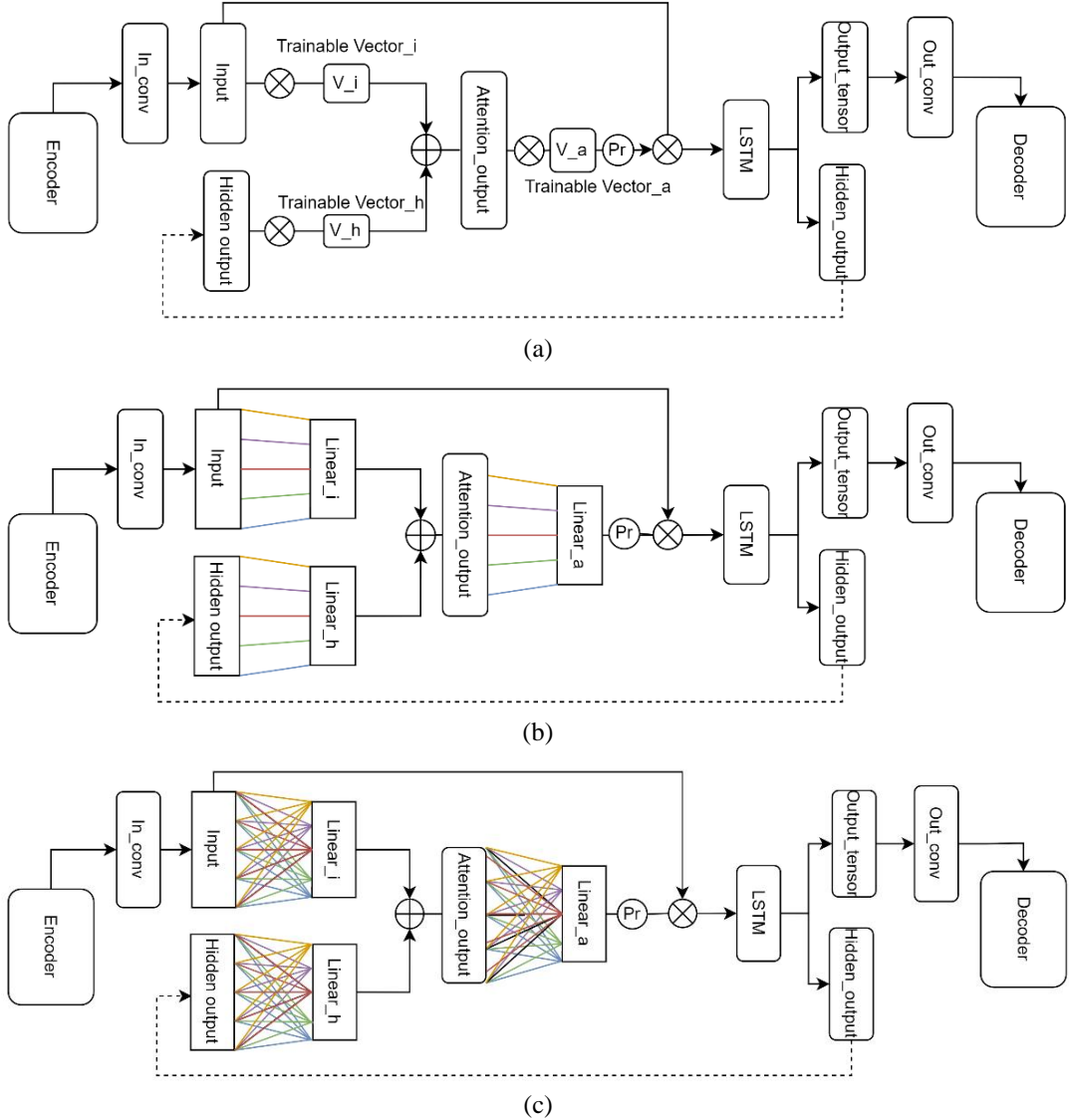

(a)

(b)

(c)

**Fig. 3.** The illustrations of (a) temporal attention module (Tem_Att); (b) spatial-temporal attention module (ST_Att); and (c) spatial-temporal attention module with fully connected layers (STFC_Att).

The structure of STFC_Att is demonstrated in **Fig. 3(c)**, where different colored lines represent the many-to-many connections between the input feature matrix, the hidden state output, and the attention output, along with their corresponding learnable weight matrices $U$, $H$, and $W$, denoted in **Fig. 3(c)** as Linear_i, Linear_h, and Linear_a, respectively. Each of these matrices is implemented as a trainable fully connected layer of size $1 \times 128$. These weight matrices dynamically adjust the importance of both spatial and temporal features, ensuring a robust and comprehensive feature extraction.

Similar to Tem_Att and ST_Att, in the STFC_Att module, the attention output $\overline{x^{(t-N+n)}}$ (calculated using **equation (4)**) is passed through a linear layer (size $1 \times 128$) with many to many connections, scaled between 0 and 1 using softmax (denoted by 'Pr' in **Fig. 3(c)**), and then transferred to the decoder via a convolutional layer.

The key distinction between ST_Att and STFC_Att lies in their ability to capture spatial dependencies. While the ST_Att module focuses on weighting local spatial features within each frame, the fully connected mechanism in STFC_Att extends the network's capability by establishing interrelations between spatial features across the entire frame and throughout the input image sequence. This enhancement allows the model to distinguish among ST features more effectively and to focus its attention on the most relevant patterns.

*C. Implementation Details*

**1) Deep neural network details**

On the whole, as illustrated in **Fig. 1**, the proposed method adopts an "encoder-attention module-decoder" based sequence-to-one architecture. UNet [24] is used as the neural network backbone, in which there are one *In_ConvBlock* and four consecutive down-sampling convolutional blocks in the encoder part, and four symmetrical upsampling convolutional blocks in the decoder part. Between the encoder and the decoder, there is the attention module with a temporal feature extractor (e.g., LSTM) embedded. **TABLE I** illustrates in detail the input and output sizes, as well as the parameters of each layer in the entire deep neural network.

**2) Loss function**

Vision-based lane detection can be considered as the pixel-wise binary classification problem, for which cross-entropy is a suitable loss function [28]. It is important to note that, in most cases, the pixels classified as "lanes" are far fewer than those classified as "not lanes" (i.e., the background), which makes it an unbalanced discriminative binary classification problem. Therefore, this study adopts the weighted cross-entropy as the loss function with two rescaling weights given to each class. The two weights for lane class and background class are set to the inverse proportion of the number of pixels in the two classes, i.e., there are fewer lane pixels than the background, so the weight of the lane class is larger. The adopted weighted binary cross-entropy loss function is illustrated by **equation (13)**.

$$Loss = -\frac{1}{M}\sum_{m=1}^{M}\left[w_l * y_m * log\left(h_\theta(x_m)\right) + w_{nl} * (1-y_m) * log\left(1 - h_\theta(x_m)\right)\right] \quad (13)$$

where $M$ is the number of training examples; $w_l$ stands for the weight of lane class, while $w_{nl}$ for the background class; $y_m$ is the true target label for training example $m$; $x_m$ is the input for training example $m$; and $h_\theta$ is the neural network model with weights $\theta$.

**3) Training details**

Various variants of the developed neural network model, as well as selected baseline models, had been trained and tested on the Dutch national high-performance supercomputer cluster Lisa using four Titan RTX GPUs with the data trained parallelly in PyTorch using *torch.nn.DataParallel()*. The input image size is set as $128 \times 256$, and the training batch size is set as 64. The learning rate is initially set to 0.01 with decay applied after each epoch. The Adam [29], RAdam optimizer [30], and Stochastic Gradient Descent (SGD) [31] optimizers were all tested. Experiments demonstrated that SGD delivered the smallest loss in this study. Thus, the SGD optimizer was chosen, and the momentum term was applied.

## III. EXPERIMENTS AND RESULTS

To verify the effectiveness and robustness of the proposed model with the designed attention module, extensive experiments were carried out on three commonly used large-scale open-sourced datasets, i.e., TuSimple (https://github.com/TuSimple/tusimple-benchmark/), tvtLANE [17], and LLAMAS [32] datasets. Several DNN based lane detection models, e.g., LaneNet [14], SCNN [13], Seg-Net [26], UNet [24], SegNet_ConvLSTM [17], and UNet_ConvLSTM [17], were selected as the baselines.

*A. Test on tvtLANE and TuSimple Datasets*

**1) Datasets description**

The original dataset of the TuSimple Lane Detection Challenge consists of 3,626 training and 2,782 testing one-second clips that are collected under different driving conditions. Each clip is extracted into 20 continuous frames, and only the last frame, i.e., the 20th frame, is labeled as the ground truth. Additionally, Zou et al. [17] added the label of the 13th frame and augmented the TuSimple dataset with 1,148 additional clips (with also the 13th and 20th frames labeled) regarding rural road driving scenes collected in China. Moreover, rotation, flip, and crop operations are employed for data augmentation, and finally, a total number of (3,626 + 1,148) × 4 = 19,096 sequences were produced, among which 38,192 frames are labeled with ground truth.

For testing, there are 2,782 testing clips in the original TuSimple dataset. While in tvtLANE, there are two different testing sets, namely, Testset #1 which is based on the original TuSimple test set for normal driving scene testing, as well as Testset #2 which contains 12 challenging driving scenarios for testing challenging scenes and assessing the model robustness.

In the training phase, three different sampling strides, with an interval of 1, 2, and 3 frames respectively, were adopted to adapt to different driving speeds which also augment the training samples by three times, whereas in the test phase, the sampling stride was set as a fixed interval of 1 frame.

Detailed descriptions of the two datasets and sampling settings can be found in [17], [18].

**2) Qualitative evaluation**

Samples of the results from lane detection segmentation on tvtLANE testset #1 (normal driving scene), tvtLANE testset #2 (challenging scenes), and TuSimple test set are shown in **Fig. 4**, **Fig. 5**, and **Fig. 6**, respectively. The lane lines are segmented into white pixels, while the background is displayed in black pixels. Three proposed attention-based model variants and the baseline deep learning models were tested. Here in Fig. 4, all of the results are without post-processing, which also applies to all the visualizations and quantitative evaluations discussed later in this paper.

Qualitatively, the models should be able to a) correctly predict the number of lanes; b) accurately locate the lane lines in the segmentation image; c) segment the lanes in thin lines without blurs; d) keep proper continuity without unexpected breaks in continuous lanes. Regarding these aspects, the proposed models with attention mechanisms all deliver good results, especially the STFC_Att-based model indicated in the last row (i), which outputs the thinner lane lines with good continuity and fewer blurs. One may argue that it does not detect the correct number of lanes in the first two columns

TABLE I

ARCHITECTURE AND LAYER-SPECIFIC PARAMETER SETTINGS OF THE NEURAL NETWORK

| Layer | | Input (channel×hight×width) | Output (channel×hight×width) | Kernel | Padding | Stride | Activation |
|---|---|---|---|---|---|---|---|
| In_ConvBlock | In_Conv_1 | 3×128×256 | 64×128×256 | 3×3 | (1,1) | 1 | ReLU |
| | In_Conv_2 | 64×128×256 | 64×128×256 | 3×3 | (1,1) | 1 | ReLU |
| Down_ConvBlock_1 | Maxpool_1 | 64×128×256 | 64×64×128 | 2×2 | (0,0) | 2 | --- |
| | Down_Conv_1_1 | 64×64×128 | 128×64×128 | 3×3 | (1,1) | 1 | ReLU |
| | Down_Conv_1_2 | 128×64×128 | 128×64×128 | 3×3 | (1,1) | 1 | ReLU |
| Down_ConvBlock_2 | Maxpool_2 | 128×64×128 | 128×32×64 | 2×2 | (0,0) | 2 | --- |
| | Down_Conv_2_1 | 128×32×64 | 256×32×64 | 3×3 | (1,1) | 1 | ReLU |
| | Down_Conv_2_2 | 256×32×64 | 256×32×64 | 3×3 | (1,1) | 1 | ReLU |
| Down_ConvBlock_3 | Maxpool_3 | 256×32×64 | 256×16×32 | 2×2 | (0,0) | 2 | --- |
| | Down_Conv_3_1 | 256×16×32 | 512×16×32 | 3×3 | (1,1) | 1 | ReLU |
| | Down_Conv_3_2 | 512×16×32 | 512×16×32 | 3×3 | (1,1) | 1 | ReLU |
| Down_ConvBlock_4 | Maxpool_4 | 512×16×32 | 512×8×16 | 2×2 | (0,0) | 2 | --- |
| | Down_Conv_4_1 | 512×8×16 | 512×8×16 | 3×3 | (1,1) | 1 | ReLU |
| | Down_Conv_4_2 | 512×8×16 | 512×8×16 | 3×3 | (1,1) | 1 | ReLU |
| Attention Module | In_Attention_Conv_5_1 | 512×8×16 | 1×8×16 | 1×1 | --- | 1 | --- |
| | AttentionLayer_1 | 1×128* | 1×128* | --- | --- | --- | --- |
| | AttentionLayer_2 | 1×128* | 1×128* | --- | --- | --- | --- |
| | AttentionLayer_3 | 1×128* | 1×128* | --- | --- | --- | --- |
| | LSTM | 128 | 128 | --- | --- | --- | --- |
| | Out_Attention_Conv_5_2 | 1×8×16 | 512×8×16 | 1×1 | --- | 1 | --- |
| Up_ConvBlock_4 | UpsamplingBilinear2D_1 | 512×8×16 | 512×16×32 | 2×2 | (0,0) | 2 | --- |
| | Up_Conv_4_1 | 1024×16×32 | 256×16×32 | 3×3 | (1,1) | 1 | ReLU |
| | Up_Conv_4_2 | 256×16×32 | 256×16×32 | 3×3 | (1,1) | 1 | ReLU |
| Up_ConvBlock_3 | UpsamplingBilinear2D_2 | 256×16×32 | 256×32×64 | 2×2 | (0,0) | 2 | --- |
| | Up_Conv_3_1 | 512×32×64 | 128×32×64 | 3×3 | (1,1) | 1 | ReLU |
| | Up_Conv_3_2 | 128×32×64 | 128×32×64 | 3×3 | (1,1) | 1 | ReLU |
| Up_ConvBlock_2 | UpsamplingBilinear2D_3 | 128×32×64 | 128×64×128 | 2×2 | (0,0) | 2 | --- |
| | Up_Conv_2_1 | 256×64×128 | 64×64×128 | 3×3 | (1,1) | 1 | ReLU |
| | Up_Conv_2_2 | 64×64×128 | 64×64×128 | 3×3 | (1,1) | 1 | ReLU |
| Up_ConvBlock_1 | UpsamplingBilinear2D_4 | 64×64×128 | 64×128×256 | 2×2 | (0,0) | 2 | --- |
| | Up_Conv_1_1 | 128×128×256 | 64×128×256 | 3×3 | (1,1) | 1 | ReLU |
| | Up_Conv_1_2 | 64×128×256 | 64×128×256 | 3×3 | (1,1) | 1 | ReLU |
| Out_ConvBlock | Out_Conv | 64×128×256 | 2×128×256 | 1×1 | (0,0) | 1 | --- |

*This is an example of the spatial-temporal attention (ST_Att) module. Corresponding to three attention variants, parameters in AttentionLayer_1, AttentionLayer_2, and AttentionLayer_3 will be learnable vectors of size 1 × 1 for Tem_Att, learnable vectors of size 1 × 128 for ST_Att, or learnable vectors with many-to-many connections of size 1 × 128 for STFC_Att, respectively.

from the left. However, when zooming in for details, one can identify that the model correctly detects the left road boundary lanes which are too difficult and not labeled in the ground truth. This defect with the dataset is also discussed in [22].

Furthermore, in accordance with previous studies [17], [18], models using multi-continuous image frames generally outperform models using a single frame, as shown in Fig. 4, the latter ones output thick lines with heavy blurs.

According to **Fig. 5**, the proposed model is compared qualitatively with the baseline models on some extremely challenging driving scenarios (tested on tvtLANE testset #2). Involving a broad range of challenging situations, testset #2 is a separate new dataset which is unseen during the training phase. It is observed that all the models do not perform well, especially regarding the 3rd column where there are vehicle occlusions and dirt road surfaces simultaneously. However, similar to norm scenes, the proposed attention-based models overall surpass baselines with thinner continuous lines and more correct locations and lane numbers. Typically, shown in the 4th column of **Fig. 5**, the STFC_Att-UNet_LSTM model demonstrates superior results in detecting smooth clear lines with the correct number of lanes in the serious vehicle occlusion case, in which almost all the other models are defeated. This can be inferred by its capability of exploring spatial-temporal correlations among neighboring pixels.

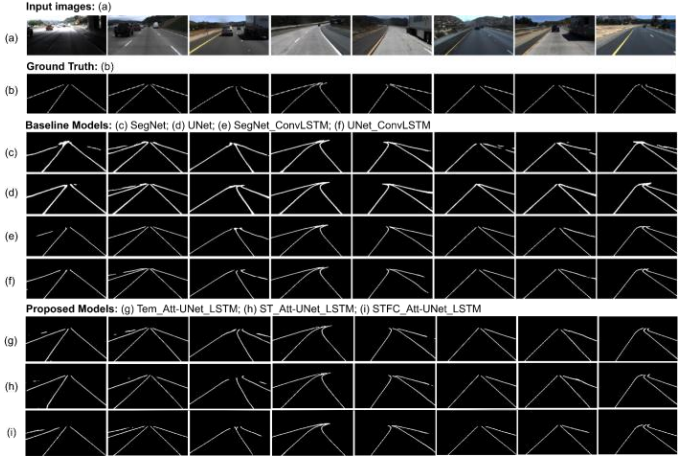

**Fig. 4.** Qualitative evaluation 1: Comparison of the lane detection results on tvtLANE testset #1 (normal scenes).

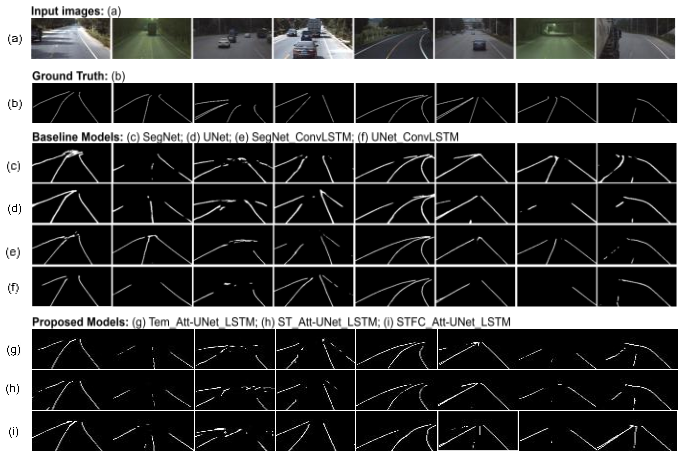

**Fig. 5.** Qualitative evaluation 2: Comparison of the lane detection results on tvtLANE testset #2 (challenging scenes).

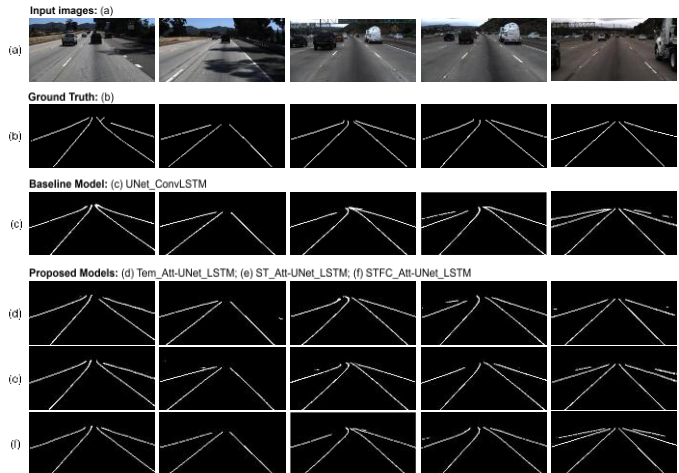

**Fig. 6.** Qualitative evaluation 3: Comparison of the lane detection results on the TuSimple test set.

The TuSimple test set is similar to the tvtLANE testset #1, thus similar patterns are observed in **Fig. 6**. Compared with the baseline model UNet_ConvLSTM, the proposed models can detect more correct lane lines with fewer blurs.

### 3) Quantitative evaluation

***Evaluation metrics***: Treating the vision-based lane detection as a pixel-wise unbalanced two-class classification and discriminative segmentation task, and following the convention in previous studies [13], [17], [18], [22], [33], [34], this study utilizes four commonly adopted evaluation criteria, i.e., accuracy, precision, recall, and F1-measure, to quantitatively verify the proposed models. The four criteria are illustrated in **equations (14)~(17)**:

$$\text{Accuracy} = \frac{Truly\ Classified\ Pixels}{Total\ Number\ of\ Pixels} \quad (14)$$

$$\text{Precision} = \frac{\text{True Positive}}{\text{True Positive+False Positive}} \quad (15)$$

$$\text{Recall} = \frac{\text{True Positive}}{\text{True Positive+False Negative}} \quad (16)$$

$$\text{F1-measure} = \frac{2*\text{Precision}*\text{Recall}}{\text{Precision+Recall}} \quad (17)$$

where true positive correlates to the pixels that are accurately identified as lanes, false-positive indicates the number of background pixels that are incorrectly categorized as lanes, and false negative is for the number of lane pixels that were incorrectly categorized as background.

This study also provides the size of the model parameter, referred to as Params (M), as well as multiply-accumulate (MAC) operations, referred to as MACs (G), as indicators for estimating the computational complexity and capabilities of the models for real-time performance.

***Quantitative comparison on tvtLANE testset #1***: As shown in **TABLE II**, when testing on tvtLANE testset#1, all the developed attention-based models perform better than the baselines regarding F1-Measure, accuracy, and precision. This verifies the effectiveness of the proposed attention mechanism. The developed model STFC_Att-SCNN_UNet_LSTM (which will be discussed in detail in the ablation study) performs the best with the highest F1-Measure, accuracy, and precision. Furthermore, compared to the other two baseline models, i.e., UNet_ConvLSTM and SegNet_ConvLSTM, which also adopt multiple frames as inputs, the developed models are all smaller in parameter size and with fewer MACs. This means that the developed models are more efficient and can deliver better results while using lower computational resources and with higher processing speed.

***Quantitative comparison on tvtLANE testset #2***: To test

TABLE II

MODEL QUANTITATIVE PERFORMANCE COMPARISON ON TVTLANE TESTSET #1 (NORMAL SCENES)

| | Models | Test_Acc (%) | Precision | Recall | F1-Measure | MACs (G) | Params (M) |
|---|---|---|---|---|---|---|---|
| | | Baseline Models | | | | | |
| Models using single image | U-Net | 96.54 | 0.790 | **0.985** | 0.877 | 15.5 | 13.4 |
| | SegNet | 96.93 | 0.796 | 0.962 | 0.871 | 50.2 | 29.4 |
| | SCNN* | 96.79 | 0.654 | 0.808 | 0.722 | 77.7 | 19.2 |
| | LaneNet* | 97.94 | 0.875 | 0.927 | 0.901 | 44.5 | 19.7 |
| Models using continuous image sequence | SegNet_ConvLSTM | 97.92 | 0.874 | 0.931 | 0.901 | 217.0 | 67.2 |
| | UNet_ConvLSTM | 98.00 | 0.857 | 0.958 | 0.904 | 69.0 | 51.1 |
| | | Proposed Models | | | | | |
| | Tem_Att-UNet_LSTM | 98.08 | 0.877 | 0.936 | 0.906 | 44.7 | 13.5 |
| | ST_Att-UNet_LSTM | 98.09 | 0.879 | 0.941 | 0.909 | 44.8 | 13.5 |
| | STFC_Att-UNet_LSTM | **98.14** | **0.887** | 0.941 | **0.911** | 44.9 | 13.5 |
| | *STFC_Att-SCNN_UNet_LSTM*** | *98.20* | *0.906* | *0.936* | *0.921* | *68.9* | *13.7* |

TABLE III

MODEL QUANTITATIVE PERFORMANCE COMPARISON ON TVTLANE TESTSET #2 (12 CHALLENGING SCENES)

| Challenging Scenes — Models | 1-curve & occlude | 2-shadow | 3-bright | 4-occlude | 5-curve | 6-dirty & occlude | 7-urban | 8-blur & curve | 9-blur | 10-shadow | 11-tunnel | 12-dim & occlude |
|---|---|---|---|---|---|---|---|---|---|---|---|---|
| | PRECISION | | | | | | | | | | | |
| U-Net | 0.7018 | 0.7441 | 0.6717 | 0.6517 | 0.7443 | 0.3994 | 0.4422 | 0.7612 | 0.8523 | 0.7881 | 0.7009 | 0.5968 |
| SegNet | 0.6810 | 0.7067 | 0.5987 | 0.5132 | 0.7738 | 0.2431 | 0.3195 | 0.6642 | 0.7091 | 0.7499 | 0.6225 | 0.6463 |
| UNet_ConvLSTM | 0.7591 | 0.8292 | 0.7971 | 0.6509 | 0.8845 | 0.4513 | **0.5148** | 0.8290 | **0.9484** | 0.9358 | 0.7926 | **0.8402** |
| SegNet_ConvLSTM | 0.8176 | 0.8020 | 0.7200 | 0.6688 | 0.8645 | **0.5724** | 0.4861 | 0.7988 | 0.8378 | 0.8832 | 0.7733 | 0.8052 |
| Tem_Att-UNet_LSTM | **0.8430** | **0.8909** | 0.7732 | 0.5740 | 0.8322 | 0.4692 | 0.4567 | **0.8358** | 0.8090 | 0.9244 | 0.7893 | 0.8046 |
| ST_Att-UNet_LSTM | 0.7938 | 0.8743 | **0.8013** | 0.7014 | **0.8894** | 0.5215 | 0.4935 | 0.8290 | 0.8517 | 0.9286 | 0.7516 | 0.8218 |
| STFC_Att-UNet_LSTM | 0.8239 | 0.8782 | 0.7646 | **0.7031** | 0.8871 | 0.5295 | 0.4848 | 0.7354 | 0.9023 | **0.9395** | **0.8794** | 0.7542 |
| | F1-MEASURE | | | | | | | | | | | |
| U-Net | 0.8200 | 0.8408 | 0.7946 | 0.7337 | 0.7827 | 0.3698 | 0.5658 | 0.8147 | 0.7715 | 0.6619 | 0.5740 | 0.4646 |
| SegNet | 0.8042 | 0.7900 | 0.7023 | 0.6127 | 0.8639 | 0.2110 | 0.4267 | 0.7396 | 0.7286 | 0.7675 | 0.6935 | 0.5822 |
| UNet_ConvLSTM | 0.8465 | **0.8891** | 0.8411 | 0.7245 | 0.8662 | 0.2417 | **0.5682** | 0.8323 | **0.7852** | 0.6404 | 0.4741 | 0.5718 |
| SegNet_ConvLSTM | 0.8852 | 0.8544 | 0.7688 | 0.6878 | **0.9069** | **0.4128** | 0.5317 | 0.7873 | 0.7575 | **0.8503** | **0.7865** | **0.7947** |
| Tem_Att-UNet_LSTM | **0.8933** | 0.8657 | 0.8123 | 0.6513 | 0.8306 | 0.3530 | 0.5263 | 0.8290 | 0.7039 | 0.5338 | 0.5225 | 0.5226 |
| ST_Att-UNet_LSTM | 0.8548 | 0.8977 | 0.8253 | 0.7293 | 0.8254 | 0.3627 | 0.5543 | **0.8369** | 0.7480 | 0.6197 | 0.5522 | 0.5363 |
| STFC_Att-UNet_LSTM | 0.8690 | 0.9059 | 0.8314 | **0.7456** | 0.8086 | 0.3660 | 0.5277 | 0.7715 | 0.7329 | 0.6543 | 0.6471 | 0.5852 |

TABLE IV

MODEL QUANTITATIVE PERFORMANCE COMPARISON ON *TUSIMPLE* TEST SET

| Models | Test_Acc (%) | Precision | Recall | F1-Measure | MACs (G) | Params (M) |
|---|---|---|---|---|---|---|
| | Baseline Models | | | | | |
| SegNet_ConvLSTM* | 97.96 | 0.852 | **0.964** | 0.901 | 217.0 | 67.2 |
| UNet_ConvLSTM* | **98.22** | 0.857 | 0.958 | 0.904 | 69.0 | 51.1 |
| UNet_DoubleConvGRU* | 98.04 | 0.875 | 0.953 | 0.912 | --- | 13.4 |
| | Proposed Models | | | | | |
| Tem_Att-UNet_LSTM | 98.05 | 0.876 | 0.923 | 0.899 | 44.7 | 13.5 |
| ST_Att-UNet_LSTM | 98.14 | 0.881 | 0.925 | 0.902 | 44.8 | 13.5 |
| STFC_Att-UNet_LSTM | *98.20* | **0.886** | 0.950 | **0.917** | 44.9 | 13.5 |

Notes: * Results reported in [22].    ** Model variant used for ablation study.

*Tem_Att-UNet_LSTM* means the temporal attention based model using the UNet_LSTM network backbone. This naming rule also applies to other models.

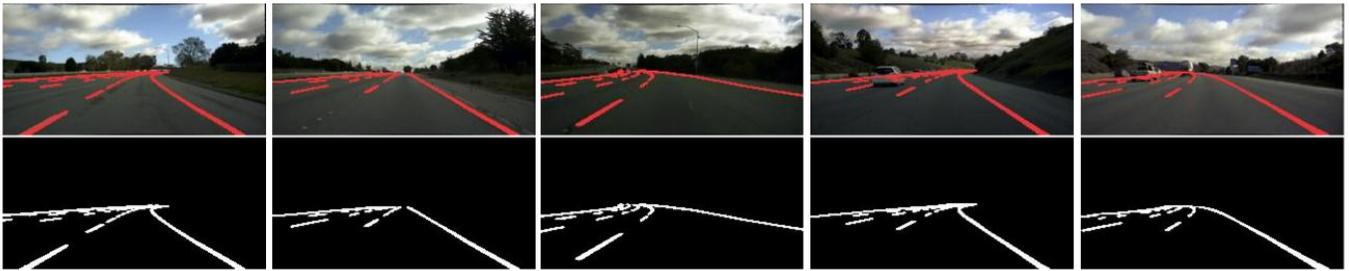

**Fig. 7.** Qualitative evaluation 4: Lane detection results on the LLAMAS dataset.

model robustness, the developed models were also evaluated and verified on the brand-new dataset, namely the tvtLANE testset #2, which contains 12 challenging scenes.

As shown in **TABLE III**, in terms of precision, ST_Att-UNet_LSTM performs the best in bright, curved, and urban scenes, while STFC_Att-UNet_LSTM performs the best in occluded, shadow, and tunnel scenes. Therefore, they dominate half of the 12 challenging scenes.

High precision means the model is more strict for the pixels to be classified as lane lines, i.e., fewer False Positives. This is crucial for the vehicles' localizing lanes. However, being too strict might result in more False Negatives, then a lower recall ratio, and a worse F1-measure. This is why the developed models are not good in terms of F1-measure. Furthermore, it was witnessed that during the training process, all the models obtained higher recalls and lower precisions at the beginning. Then, as the training went on, the recalls decreased while the precisions rose. This general pattern applies to all models. With this, one can infer that a higher precision is more important. All these demonstrate the developed models' robustness over challenging scenes.

***Quantitative comparison on TuSimple test***: The TuSimple test set has similar but more testing samples compared to the tvtLANE testset#1. Regarding the quantitative results on the TuSimple test set, as demonstrated in **TABLE IV**, the proposed STFC_Att-UNet_LSTM obtains the best F1-measure, the best precision, and the second-best accuracy (i.e., 98.20%, only a bit lower than the best of 98.22%). Although UNet_ConvLSTM shows the best accuracy, it is worth noting that its MACs and parameter size are much larger than those of the proposed models. In this case, one can conclude that the developed models with lower computational complexities are more efficient and robust with competitive results on the TuSimple test set.

### B. Test on LLAMAS dataset

#### 1) Datasets description

To further verify the robustness of the proposed method, the LLAMAS dataset [32] is adopted to train, validate, and test different models. Consisting of a total of 100,042 images, LLAMAS is one of the largest open-source lane marker datasets. Among the 100,042 images, 79,113 of them are used for training with labeled ground truth, while 20,929 of them were originally used for testing with no corresponding labels.

To still follow the proposed end-to-end supervised learning pipeline and make it comparable with the previous work [22], this study follows the processes described in [22] utilizing only the labeled 79,113 images and dividing them into two groups. To be specific, 58,269 images were used for training, and 20,844 images were used for testing. More details about the LLAMAS dataset can be found in [22], [32].

#### 2) Qualitative evaluation

Limited by computational resources and time, this study only trained ST_Att-UNet_LSTM and STFC_Att-UNet_LSTM models on the LLAMAS dataset.

**Fig. 7** provides qualitative visualization results of ST_Att-UNet_LSTM for testing on the LLAMAS dataset. In the top row, the predicted lane lines are shown in red color, and in the bottom row, the predicted lane lines are segmented with white pixels under black background. As shown, the lane lines in LLAMAS are labeled in a different way using dash lines, which makes it much more challenging. Qualitatively, from the visualization, it is observed that there are very few false positives and lane lines are generally predicted accurately.

#### 3) Quantitative evaluation

To quantitatively evaluate the model performances on the LLAMAS dataset, except for the aforementioned precision and recall, similar to [22], [32], mean Average Precision (mAP) was also adopted. mAP is defined as the arithmetic mean of the Average Precision (AP) scores calculated for each individual image frame. For a single image, the AP represents the area under the Precision-Recall curve, approximated as the sum of weighted precision scores where the weights are the increments in recall between successive thresholds. To be clear, mAP is illustrated in **equation (18)**:

$$mAP = 1/T \sum_{p=1}^{T} (\sum_{q=1}^{V+1} Precision_{p,q} * \Delta Recall_{p,q}) \quad (18)$$

where $T$ means the total number of the tested image frames; $V$ means number of recall thresholds per image; $\Delta Recall_{p,q}$ is the difference between Recall values of two consecutive samples for frame $p$; $Precision_{p,q}$ is the Precision at the $q$-th threshold for frame $p$. In the implementation, similar to [22], this study sets $Recall_0$ to 0, $Precision_0$ to 1, and $V$ to 100.

The quantitative results are demonstrated in **TABLE V**. As shown in **TABLE V**, the STFC_Att-UNet_LSTM model provides the best corner precision when testing on the LLAMAS dataset. This is an indication that the model delivers a lower number of false positives, which, as discussed before,

| Models | Average Precision (AP) | Precision | Recall |
|---|---|---|---|
| UNet_Double_ConvGRU* | **0.8519** | 0.6162 | 0.6163 |
| SegNet ConvLSTM* | 0.8500 | 0.5487 | **0.6839** |
| UNet_ConvLSTM* | 0.8510 | 0.5857 | 0.6558 |
| ST_Att-UNet_LSTM | 0.7106 | 0.6253 | 0.6584 |
| STFC_Att-UNet_LSTM | 0.7141 | **0.6317** | 0.6413 |

Notes: * Results reported in [22].

is more crucial for lane localization. It also obtains a comparable corner recall. Furthermore, it is worth noting that both the proposed models maintain a better balance among the three evaluation metrics, although they do not perform well in average precision. To sum up, the developed models' robustness on the LLAMAS dataset is demonstrated with competitive quantitative and qualitative detection results.

### C. Qualitative Test on Unlabeled Netherlands Lane Dataset

To further verify the developed models' robustness in handling new and challenging driving scenes, the unlabeled Netherlands lane dataset was adopted for qualitative testing. This dataset covers a wide range of driving situations in the Netherlands, some of which are very challenging.

**Fig. 8** shows the lane detection results of ST_Att-UNet_LSTM, which is only trained on the LLAMAS dataset. Even without any supervised training on the unlabeled Netherlands lane dataset, the proposed model demonstrates excellent transfer capabilities by clearly detecting lane line numbers and locations. Furthermore, the model can correctly identify whether the lanes are continuous or dashed lines. The good performance can be attributed to that the developed ST_Att-UNet_LSTM with spatial-temporal attention module can aggregate rich valuable context information to focus on generalized information and salient regions in both one image and the continuous image frames. This qualitative testing further verifies the robustness of the developed model.

### IV. ABLATION STUDY AND DISCUSSION

#### A. Post-explanation of Attention Mechanism by Visualization

To elucidate the functionality of the proposed spatial-temporal attention mechanism, this subsection presents a case study using feature map visualizations. Consider a scenario where a vehicle is traveling under a bridge, as depicted in **Fig. 9**, in this instance, shadows cast by the bridge obscure portions of the road, rendering the lane markings indiscernible (even to human observers). Furthermore, on the right side, lane markings are partially occluded by a preceding vehicle.

The top row (a) displays the original sequence of continuous image frames, illustrating the vehicle's gradual movement beneath the bridge from frame 1 to frame 5 (left to right). Rows (b), (c), and (d) compare the feature map activations at *Up_ConvBlock_4* (the first upsampling block as shown in **TABLE I**) for UNet, UNet-ConvLSTM, and STFC_Att_UNet_LSTM, respectively. Since all three models share the *Up_ConvBlock_4* structure, which immediately follows the attention module in STFC_Att_UNet_LSTM and the ConvLSTM module in UNet-ConvLSTM, this comparison provides a meaningful evaluation of their performance.

The baseline UNet model exhibits strong activation primarily along the leftmost lane, while detection on the right appears fragmented. Critically, in distant regions, the detected left and right lanes converge erroneously, accompanied by blurred and spurious activations. This limitation reflects the model's insufficient contextual reasoning for maintaining coherent lane structures in occluded and distant areas. By integrating ConvLSTM, UNet_ConvLSTM demonstrates an improved ability to detect lane markings in occluded regions through temporal dependency modeling. However, its overall activation intensity remains subdued relative to the baseline UNet, suggesting constraints in spatial feature extraction and suboptimal spatial-temporal correlation. In contrast, STFC_Att_UNet_LSTM outperforms both models by maintaining consistent, non-converging activation patterns, particularly excelling in scenarios with partial or complete occlusions. Its spatial-temporal attention mechanism can dynamically weigh the importance of the frame based on lane visibility, enabling robust inference of lane positions even in challenging scenarios. The enhanced performance is attributed to the model's capacity to establish strong interrelations among spatial features across sequential frames, effectively "memorizing" lane positions from previous observations.

Ultimately, the spatial-temporal attention-based neural network leverages information from prior frames to predict lane locations, even when they are entirely obscured in the current frame (e.g., the $5^{th}$ frame in row (a)). As shown in frames 1–4, lane markings are partially visible, allowing the model to detect salient regions of interest (highlighted by pronounced bright activations in **Fig. 9**). By leveraging this accumulated ST information, the model accurately predicts lane positions in frame 5. This ability to retain and utilize learned ST correlations across frames enhances robustness in adverse driving conditions.

#### B. Comparisons between Three Model Variants

Comparing the three proposed model variants' results in **Tables II**, **III**, **IV**, and **V**, it is demonstrated that STFC_Att-UNet_LSTM outperforms Tem_Att-UNet_LSTM and ST_Att-UNet_LSTM in various situations and regarding different metrics; while ST_Att-UNet_LSTM is also generally better than Tem_Att-UNet_LSTM. This can be explained by the fact that Tem_Att-UNet_LSTM only gets the temporal attention mechanism which does not consider the interrelationship among the pixels and different regions; while, in the ST_Att-UNet_LSTM, with the one-to-one connection, it can learn the importance of the individual pixel with the weights and hidden layer but without the knowledge of neighboring pixels; and finally, in the STFC_Att-UNet_LSTM, using the many-to-many connection, the spatial dependencies between the pixels are incorporated, along with the temporal correlations among the continuous frames. Thus, the STFC_Att-UNet_LSTM is the real "spatial-temporal" attention, and in this way, the verification of the strengths of the proposed spatial-temporal attention mechanism is further enhanced.

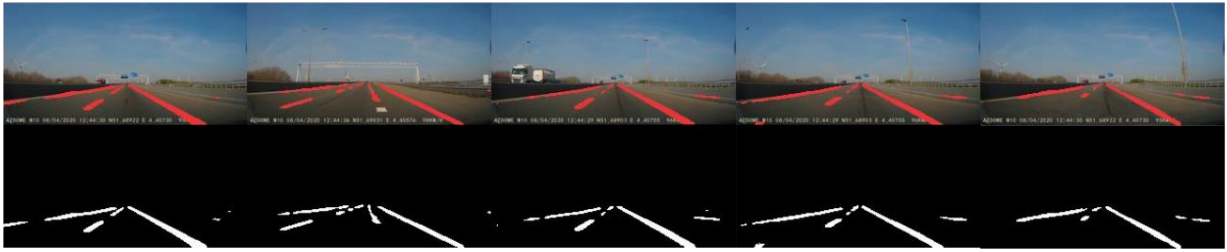

**Fig. 8.** Qualitative evaluation 5: Lane detection results on unlabeled Netherlands lane dataset.

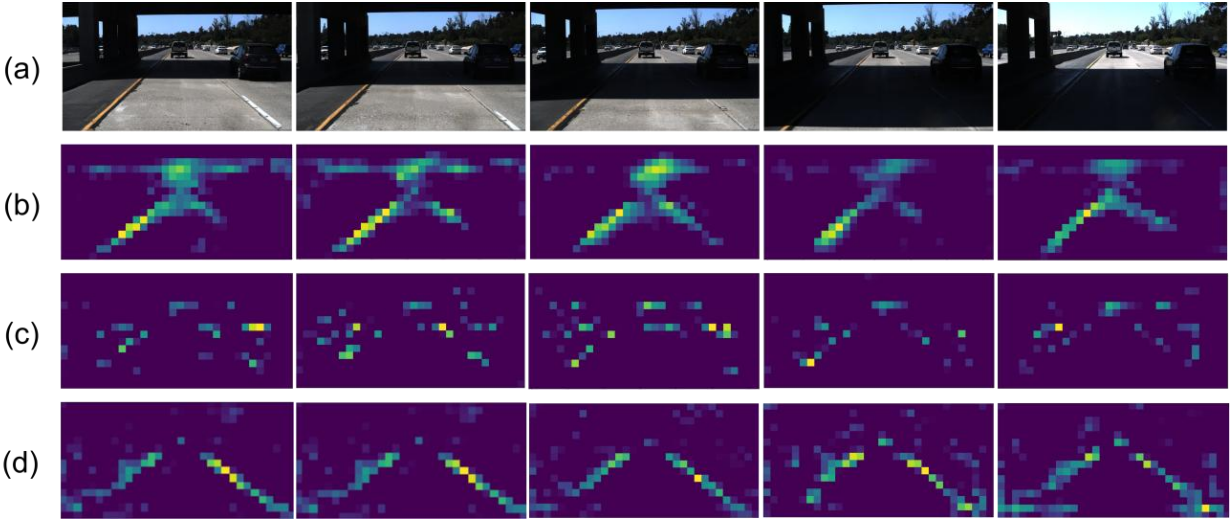

**Fig. 9.** Post-explanation visualization of the case study - under a bridge with shadows and occlusion: (a) input images; and feature map visualizations of (b) UNet, (c) UNet_ConvLSTM, (d) STFC_Att_UNet_LSTM.

### C. Cooperation with Other Model Structures and Methods

This study also investigated the compatibility of the proposed model working with other mechanisms, such as incorporating the SCNN layer to further enhance feature extraction and spatial correlation within individual images [13], [18]. The last row of **TABLE II** shows the results for STFC_Att-SCNN_UNet_LSTM**, which incorporates the SCNN layer into the proposed spatial-temporal attention mechanism. Compared to all other models, including those with the attention mechanism but without SCNN layers, STFC_Att-SCNN_UNet_LSTM** achieves the highest accuracy, precision, and F1-measure, with only minor increases in parameter size and slight increases in MACs. These findings demonstrate that embedding SCNN layers further strengthens the model's performance, confirming the compatibility of the proposed spatial-temporal attention mechanism. As the developed spatial-temporal is modular in nature, it should be able to cooperate with any other mechanisms. Additionally, the proposed encoder-decoder-based pipeline allows all the developed models to integrate seamlessly with other methodologies, such as self-supervised pre-training approaches. In particular, the employment of pre-training using the masked sequential autoencoders [23] was shown to significantly improve the performance of the STFC_Att-SCNN_UNet_LSTM model. Incorporating self-supervised pre-training not only enhances the model's accuracy but also reduces the total training time [23], further demonstrating the flexibility and adaptability of the proposed approach.

### D. Model Size and Real-time Capability

As illustrated in **TABLE II**, all the developed models with the proposed attention mechanism possess fewer parameters and lower MACs compared with the baseline SegNet_ConvLSTM and UNet_ConvLSTM, which also use continuous frames as input. Fewer parameters and lower MACs mean the models get better performance regarding processing time and real-time capability, which would be advantageous when deployed in real-world applications.

Within the developed model variants, Tem_Att-UNet_LSTM, ST_Att-UNet_LSTM, and STFC_Att-UNet_LSTM have nearly identical parameter sizes (the little difference can not be visible in one decimal), while ST_Att-UNet_LSTM and STFC_Att-UNet_LSTM get slightly larger

MACs, with STFC_Att-UNet_LSTM getting the largest among the three variants. These variations arise from differences in model architecture. Within the group of the developed models, as the MACs increase, the model's performance generally gets better (demonstrated in **Tables II, III, IV**, and **V**), making the slight computational cost increase acceptable.

The results from all these ablation experiments provide robust evidence of the effectiveness and reliability of the proposed spatial-temporal attention mechanism for lane detection. The mechanism strikes a balance between model complexity and real-time capability, ensuring practical viability in real-world diverse driving scenarios.

## V. CONCLUSION

Previous vision-based methods for lane detection often fail to account for critical image regions and their spatial-temporal (ST) salience across continuous frames, leading to poor performance under challenging driving scenarios, espeially in mixed traffic conditions. In this study, a novel spatial-temporal attention mechanism embedded within a hybrid sequence-to-one encoder-decoder neural network architecture is proposed and implemented for accurate and robust lane detection in a variety of normal and challenging driving scenarios. The proposed spatial-temporal attention mechanism can focus on key features of lane lines and exploit salient spatial-temporal correlations among continuous frames to enhance the accuracy and robustness of lane detection. Extensive experiments conducted on three large-scale open-source datasets demonstrate the robustness and superiority of the proposed model, outperforming available state-of-the-art methods in various testing scenarios. In addition, ablation studies confirm the developed spatial-temporal attention mechanism's capabilities of cooperating with other architectures and model mechanisms. Last but not least, the sequential neural network models implemented by the proposed spatial-temporal attention mechanism possess fewer parameters and smaller multiply-accumulate operations compared with other sequential baseline models, highlighting their computational efficiency.

However, it is observed that the proposed models struggle in certain challenging cases. These cases are underrepresented in the training datasets and, in some instances, include mislabeled ground truth data (as noted in [22]), hindering the models' ability to learn their patterns. Furthermore, initial tests of the models' transferability between datasets revealed that the models trained on tvtLANE underperformed on the LLAMAS dataset due to differences in lane structures and labeling formats, and vice versa. In contrast, models trained on the LLAMAS dataset performed well on the unlabeled Netherlands lane dataset due to similar lane structures.

For real-world deployment of lane detection algorithms, adaptability to diverse lane structures and types across different countries and regions is crucial. With these findings in mind, it is recommended that future research should focus on building integrated, comprehensive lane datasets and exploring domain adaptation and transfer learning methods to improve lane detection performance across diverse datasets and driving conditions.

AUTHORSHIP CONTRIBUTION STATEMENT

**Sandeep Patil**: Conceptualization, Methodology, Data curation, Formal analysis, Visualization, Validation, Writing - original draft. **Yongqi Dong**: Conceptualization, Investigation, Methodology, Data curation, Formal analysis, Visualization, Validation, Writing - original draft, Writing - review & editing. **Haneen Farah**: Supervision, Validation, Resources, Writing - review & editing, Project administration, Funding acquisition. **Hans Hellendoorn**: Supervision, Validation.

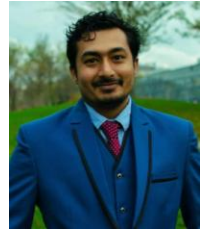

**Sandeep Patil** received the B.E. degree in mechanical engineering from People's Education Society Institute of Technology (PESIT) (autonomous), Bengaluru, India, in 2017 and the M.Sc. degree in mechanical engineering with a specialization in vehicle engineering - perception modeling, from Delft University of Technology, Delft, the Netherlands, in 2021. He is currently a Data & AI Engineer at Capgemini within the hybrid intelligence team. His research interests include deep learning, computer vision, and generative AI.

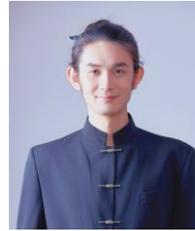

**Yongqi Dong** received the M.S. degree in control science and engineering from Tsinghua University, Beijing, China, in 2017, and the Ph.D. degree in transport and planning from Delft University of Technology, Delft, the Netherlands, in 2025. He is currently working as a Research Group Leader with the Institute of Highway Engineering, RWTH Aachen University. His research interests include deep learning, automated driving, and traffic safety. He seeks to employ AI and interdisciplinary research as tools to shape a better world.

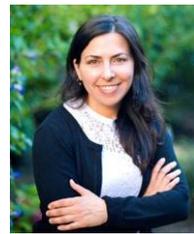

**Haneen Farah** is an Associate Professor at the Department of Transport and Planning and the Co-Director of the Traffic and Transportation Safety Lab, Delft University of Technology. She received her Ph.D. degree in transportation engineering from the Technion-Israel Institute of Technology. Before joining TU Delft, she was a Post-Doctoral Researcher at KTH Royal Institute of Technology, Stockholm, Sweden. Her research interests include road infrastructure design, road user behavior, and traffic safety. She is currently examining the implications of advances in vehicle technology and automation on infrastructure design and road user behavior.

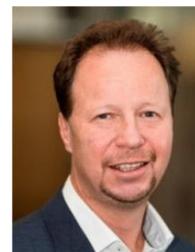

**Hans Hellendoorn** received the Ph.D. degree in computer science from the Delft University of Technology, Delft, the Netherlands, in 1990. Until 2008, he was with Siemens Research, in Germany and the Netherlands. From 1999 to 2008, he was a part-time Professor of industrial applications of computational intelligence. Since 2008, he has been a full-time Professor of control theory. From 2012 to 2018, he was the Chair of the Delft Center for Systems and Control; since April 2018, he has been Chair of the Cognitive Robotics Department. Since 2024, he has been Vice President Education at TU Delft. His research focuses on multiagent control of large-scale hybrid systems. He is the co-author of 4 scientific books and author and coauthor of more than 200 scientific publications.